\documentclass[letterpaper, 10 pt, conference]{ieeeconf} \IEEEoverridecommandlockouts
\let\labelindent\relax

\makeatletter
\def\endthebibliography{%
  \def\@noitemerr{\@latex@warning{Empty `thebibliography' environment}}%
  \endlist
}
\makeatother

\usepackage{cite}
\usepackage{amsmath,amssymb,amsfonts}
\usepackage{graphicx, import}
\usepackage{textcomp}
\usepackage{enumitem}
\usepackage{xcolor}
\usepackage{tikz}
\usepackage{verbatim}
\usetikzlibrary{shapes.gates.logic.US,trees,positioning,arrows}
\usepackage{todonotes}
\usepackage{algorithmicx}
\usepackage[ruled]{algorithm}
\usepackage{algpseudocode}

\usepackage{multirow}

\usepackage{caption}
\usepackage{subcaption}
\DeclareCaptionLabelSeparator{periodspace}{.\quad}
\graphicspath{{./images/}}

\def\BibTeX{{\rm B\kern-.05em{\sc i\kern-.025em b}\kern-.08em
    T\kern-.1667em\lower.7ex\hbox{E}\kern-.125emX}}

\usepackage{amsthm}
\newtheorem{definition}{Definition}

\begin{document}

\title{\huge Active Perception for Ambiguous Objects Classification 
\thanks{
$^1$The authors are with Istituto Italiano di Tecnologia, Genova, Italy. \newline
$^2$are also with Department of Informatics, Bioengineering, Robotics and Systems Engineering, Universit\`a di Genova, Genova, Italy
\texttt{evgenii.safronov@iit.it}
}
}

\author{{Evgenii Safronov$^{1,2}$, Nicola Piga$^{1,2}$, Michele Colledanchise$^1$, and Lorenzo Natale$^1$}}

\newcommand{\michele}[1]{\todo[inline]{#1}}
\newcommand{\nicola}[1]{\textcolor{orange}{{[\textbf{Nicola}: #1]}}}
\newcommand{\nicolarefine}[1]{\textcolor{red}{{[\textbf{Nicola}: #1]}}}
\newcommand{\mnt}{\emptyset}
\newcommand{\msR}{\mathbb{R}}
\newcommand{\msS}{\mathbb{S}}
\newcommand{\msF}{\mathbb{F}}

\newcommand{\nt}{$\emptyset$}
\newcommand{\sR}{$\mathbb{R}$}
\newcommand{\sS}{$\mathbb{S}$}
\newcommand{\sF}{$\mathbb{F}$}

\maketitle

\begin{abstract}
Recent visual pose estimation and tracking solutions provide notable results on popular datasets such as T-LESS and YCB. However, in the real world, we can find ambiguous objects that do not allow exact classification and detection from a single view.
In this work, we propose a framework that, given a single view of an object, provides the coordinates of a next viewpoint to discriminate the object against similar ones, if any, and eliminates ambiguities. We also describe a complete pipeline from a real object's scans to the viewpoint selection and classification.
We validate our approach with a Franka Emika Panda robot and common household objects featured with ambiguities. We released the source code to reproduce our experiments.
\end{abstract}

\section{Introduction}
\label{sec:intro}

Object recognition remains a fundamental skill for robots
that operate in complex environments. For example, consider the task ``fetch an apple''; this simple task requires first detecting and classifying an apple among other fruits. 
The task becomes more problematic when we ask the robot to ``fetch a ripe apple'', which requires the robot to examine the apple from different views to assess the ripeness.

%

\begin{figure}
\includegraphics[width=0.96\columnwidth]{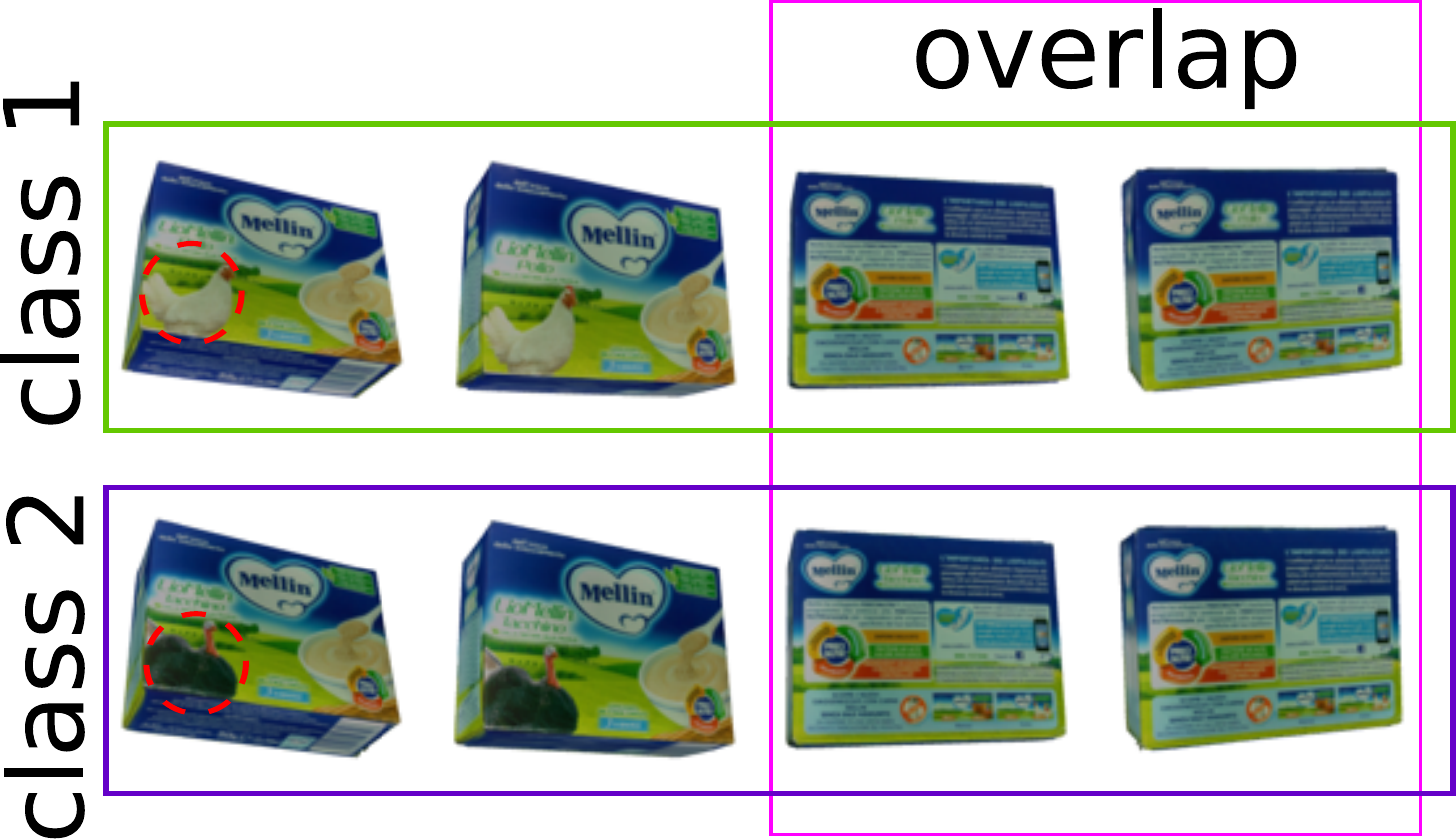}
        \caption{Example of objects visually identical except for a single side. 
        }
         \label{IN.fig.babyfood.training}
         
         \end{figure}

Several objects, especially those used in the packaging of goods, may have a distinguishing feature only on one side (e.g., paint color or instant noodles flavor).
Fig. \ref{IN.fig.babyfood.training} shows an example of food boxes that show the flavor on one side only while they look identical from the other sides.  In this case, a robot must classify objects between the different flavors; unfortunately, this may be impossible when the objects face visually identical sides (e.g., when they are stored on a shelf).
Object classification from visual data often employs neural network-based systems trained on public image datasets. However,  public datasets do not include objects visually identical except for a single side. Hence, different ambiguous objects may fall into the same class. Thus, we must create a specific dataset (e.g., employing a 3D scanner) to train a classifier for our purpose.

Consider again the example in Fig.~\ref{IN.fig.babyfood.training}. An additional problem is that the training data will include similar input images with different labels (i.e., the backside of the boxes are visually identical), jeopardizing the convergence of the training. The considerations above led to the following research questions:

\begin{itemize}
\item  \textbf{Question 1} \emph{In presence of ambiguous objects, should we exclude the non-unique views from the training data? How to exclude them correctly and automatically?} Training on the whole dataset might not converge due to different labels on identical inputs. If the images are not identical but very similar, the classifiers might overfit and not generalize correctly to the test data.

\item  \textbf{Question 2} C\emph{an actively planning for different viewpoints (e.g., a robot that moves the camera) improve ambiguous objects' classification? How to implement such an approach?} Even assuming that we can correctly train a classifier, the robot camera might face the ambiguous part of the object. If the robot was able to determine that this view would lead to unreliable classification, it could trigger an appropriate fallback behavior. For example, the robot could look at the other side of the object to acquire a non-ambiguous, classifiable view. For this purpose, we would like to estimate, given an object, the next best view for its classification.

\end{itemize}
\begin{figure}
\includegraphics[width=0.95\columnwidth]{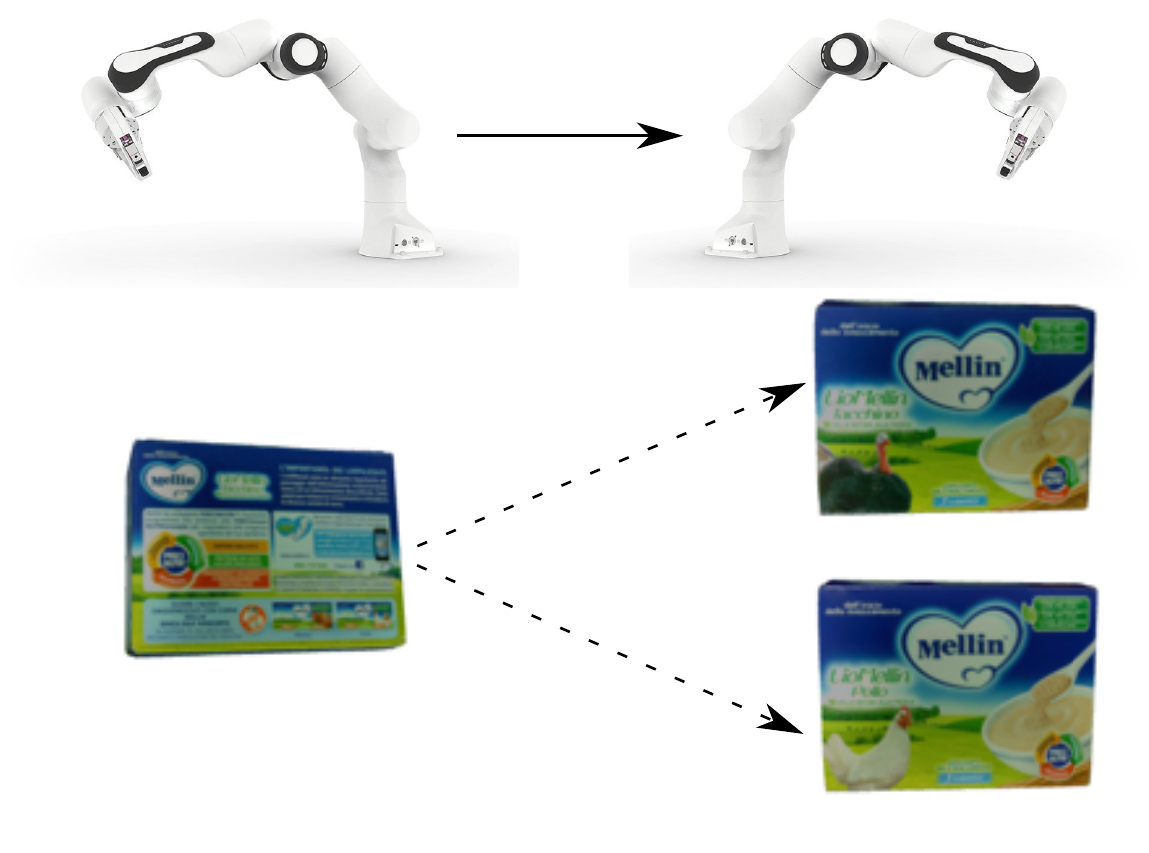}
        \caption{Possible observed views of the object after changing robot's viewpoint. While the back of both objects is similar, front sides are different enough for classification. 
        }
         \label{IN.fig.babyfood.active}
         
\vspace{-1em}
\end{figure}
In this paper, we address the above questions by proposing a novel active perception pipeline. We propose an image similarity metric based on the embedding of a denoising autoencoder \cite{Sundermeyer_2018_ECCV} for image reconstruction. The metric is used to estimate the ambiguity of object orientations and solve the problem of ambiguous objects classification.
%
Using this ambiguity score, we are able to correctly train classifiers by excluding ambiguous views from the training data. 
We propose an active perception framework that selects the next best viewpoint to minimize ambiguity and disambiguate objects using the trained classifier.

%
\section{Background}
\label{SEC.Background}
\subsection{Object classification}
Object classification is one of the most common tasks in computer vision. Machine learning approaches turned object classification into a data-oriented problem. The community is continuously improving image classification~\cite{tan2020efficientnet, pham2021meta} on standard benchmarks such as ImageNet dataset~\cite{5206848}, many works are based on popular neural networks such as ResNet~\cite{he2015deep}.  A typical training dataset for image classification consists of labeled images. It is common to use a network whose parameters have been obtained by training the network on a large, public dataset such as ImageNet, and then \emph{fine-tune} it using additional data from the task of interest. The training process is usually terminated by achieving a certain classification performance (e.g., in terms of classification accuracy) on a validation dataset that differs from the training one. If objects views are annotated with the pose of the camera, it is possible to cast the pose estimation problem as a classification or regression problems~\cite{7139363}.

\subsection{Object pose estimation with Augmented Autoencoder}
\label{SEC.Background.Autoencoder}
Deep Convolutional Neural Networks (CNNs) were successfully employed for 6D object pose estimation by regressing 2D keypoints on the object used to estimate the 6D pose via Perspective-n-Point (PNP) techniques~\cite{he2020pvn3d}. Other techniques relied on denoising autoencoders for image reconstruction in order to extract from RGB images a meaningful latent representation, conditioned on the object orientation, as in Augmented Autoencoders (AAE) \cite{Sundermeyer_2018_ECCV}.

\begin{figure}[ht!]
    \centering
    \includegraphics[width=0.99\columnwidth]{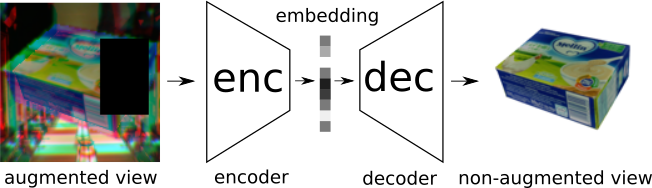}
    \caption{The training schema for AAE. Denosing autoencoder is aimed to reconstructed the non-augmented rendered view of the same as augmented input orientation.}
    \label{FIG.AAE.training}
\end{figure}

\begin{figure}[ht!]
    \centering
    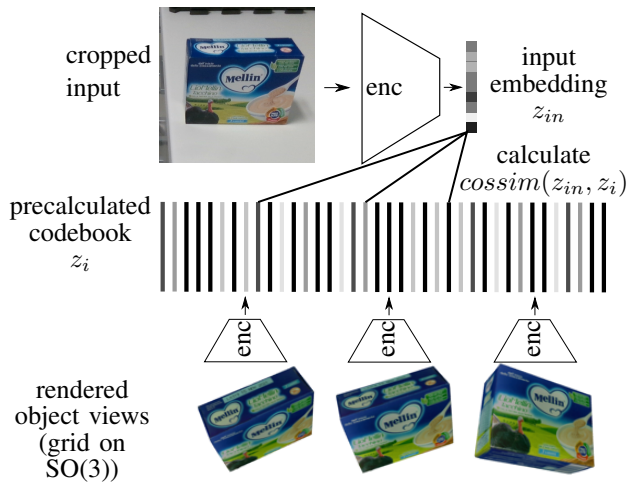
    \caption{Pose estimation with AAE. The precalculated and saved codebook of embeddings is used First we create embeedding $z_in$ of the input image}
    \label{FIG.AAE.pose_estimation}
    
\vspace{-1em}
\end{figure}

 In this context, a \emph{denoising autoencoder} is trained to reconstruct a clean rendered $view(R)$ of the object having orientation $R$ starting from an image of the same view augmented with different light conditions, noise, and small occlusions as $aug(view)$ (see Fig. \ref{FIG.AAE.training}). The autoencoder is based on CNNs and is trained to minimize the error in pixel space between the augmented and non-augmented images:
\begin{equation}
\min_{enc, dec} \sum_{i, j} ||dec(enc(aug_j(view(R_i)))) - view(R_i)||
\end{equation}
After the training procedure, the approach stores an array of embeddings generated by encoding the rendered views for uniformly sampled orientations. Such an array is called \emph{codebook}. To estimate the object orientation $R^*_{obj}$ given an input image $I$ at runtime (see Fig. \ref{FIG.AAE.pose_estimation}), one should first pass the image $I$ through the encoder of the AAE  to calculate the image embedding $z_I$. Then, find the codebook's closest entry by applying the cosine similarity between codebook and image embeddings, i.e.:
\begin{equation}
R^*_{obj} = \min_i cossim(z_{I}, z(R_i)), R_i \in \{R\}_{codebook}
\end{equation}
\begin{equation}
cossim(z_{in}, z_i) := \frac{(z_{in}, z_i)}{|z_{in}| \cdot |z_i|}
\end{equation}
The autoencoder is trained on rendered data with 3D models of objects. 

\subsection{Active perception}
Animals often act to gather information about the world through an \emph{active perception} approach \cite{gibson1966senses}. This approach, applied to robotics, was employed to improve mobile robot localization~\cite{miccol2020actperceive}, and for better object tracking in data acquisition~\cite{8246973}. Active vision is also connected to multi-view classification or pose estimation. Recent works~\cite{9340771} show that multi-view solutions improve the results over single-view ones. Other works~\cite{amb2018} employed active vision for object classification, but their active vision approach was limited to fusing the data from several fixed camera viewpoints in a passive fashion.  Active perception helps to cope with environmental uncertainties when we can move the camera viewpoint. However, correct motion planning and data interpretation might still be challenging.
\section{Approach Overview}
\label{SEC.Overview}
\begin{figure*}[ht!]
    \centering
    \includegraphics[width=1.99\columnwidth]{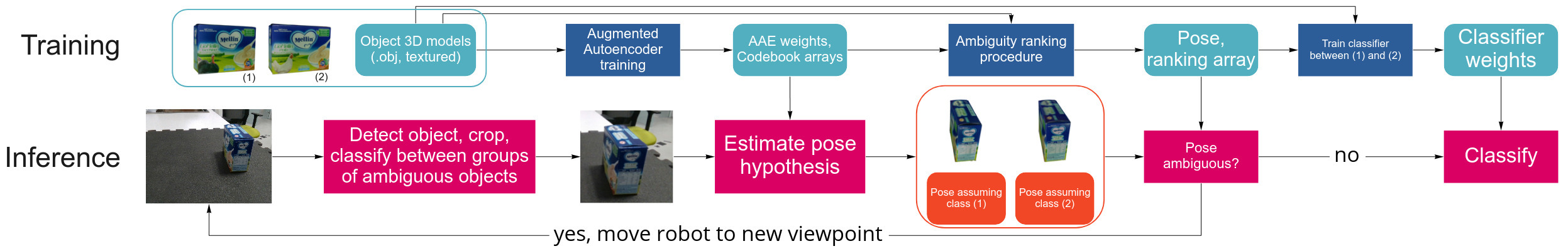}
    \caption{Proposed pipeline divided into training and inference parts.}
    \label{FIG.pipeline}
    
\vspace{-1em}
\end{figure*}

In this section, we formalize the problem and briefly describe the developed solution. 

We first address how to identify ambiguous orientations of objects. Given a pair of objects A and B, we show how to select views from object A that are the most discriminative against object B, and train a classifier solely on such views. For this purpose, we define an  \emph{ambiguity rank} to compute the next best view, which will be then used to compute the next best view to take.

Let $view^A_{R_i}$ be a view of the object $A$ that has an orientation $R$ with respect to the camera frame; the ambiguity of the object $A$ at orientation $R_i$ stems from the  existence of
an object $B$  such that, at orientation $R_j$, is  ``visually similar'' to $A$ at $R_i$. This definition implies a certain metric of image similarity $ sim(view^A_{R_i}, view^B_{R_j} )> 0$. In Section~\ref{SEC.rank} we describe in detail this similarity metric.

Note that the images would be different if taken under different light conditions, background, or due to noise. We assume that the value of the similarity between views under different rotations $R_0$ and $R_1$ is mostly unchanged in different light conditions. In other words, we assume that inequalities like $sim(view^A_{R_0}, view^B_{R}) \lessgtr sim(view^A_{R_1}, view^B_{R})$ remains true for different illuminations, therefore we can fix certain light conditions while calculating image similarity.

\begin{definition}
\label{def:ambor}
Given two objects $A,B$, the ambiguity of orientation $R_i$ of object $A$ is defined as the maximum value of a given similarity metric $sim$ between $view^A_{R_i}$ and views from object $B$:
\begin{equation}
ambiguity(R_i, A) := \max_j(sim(view^A_{R_i}, view^B_{R_j}))
\label{EQ.rank.two}
\end{equation}
\end{definition}
The most suitable object orientation that allows unambiguously classifying objects $A$ and $B$ can be then found as the solution of the following minimization:
\begin{multline}
R^A_{best} := \min_i(ambiguity(R_i, A)) =\\
 \min_i(\max_j(sim(view^A_{R_i}, view^B_{R_j})))
\label{EQ.rank.best}
\end{multline}
In case we have more than two ambiguous objects ${\mathcal{A}: A_0, A_1, A_2, ...}$,  we can define the ambiguity of the orientation  $R_i$ from the object $A_o$:
\begin{equation}
ambiguity(R_i, A_o) := \max_{k \neq o,j}(sim(view^{A_o}_{R_i}, view^{A_k}_{R_j}))
\label{EQ.rank.many}
\end{equation}
In this paper, we assume that we have a pair of ambiguous objects for simplicity. Eq.~\ref{EQ.rank.many} above allows to trivially extend the developed solutions to a group of three or more objects. We found it easier to analyze and use ambiguity if it is in the range of $[0, 1]$. We linearly transform ambiguity for each object to this range.
Now we can define ambiguous and non-ambiguous views. 
\begin{definition}
An orientation $R$ of the object $A$ from the group of ambiguous objects $\mathcal{A}$ is ambiguous if $ambiguity(R, A) \geq a$ w.r.t to chosen threshold $a$. It is non-ambiguous otherwise.
\label{DEF.amb-separation}
\end{definition}
\label{ambiguity-subset}
A lower similarity metric between two objects should result in lower ambiguity for the views at the orientations that show discriminative visual features (e.g. left hand side on Fig. \ref{IN.fig.babyfood.training}), and higher ambiguity for those that are difficult to discriminate (e.g. right hand side on Fig. \ref{IN.fig.babyfood.training}). 

We now show how to use the ambiguity to select discriminative views and train a classifier that effectively disambiguate between two objects A and B provided an image containing an object. We take a uniformly distributed subset $\{R\} \subset SO(3)$. Then, we should choose a threshold $a$ to form $\{R\}_{train}$, such that $R_i \in \{R\}_{train} \Leftrightarrow R_i \in \{R\}, ambiguity(R_i, A) \leq a$. We can sort a set of object orientations $\{R\} \subset SO(3)$ by the  ambiguity values in ascending order (as in Fig. \ref{IN.fig.tops}) to visually verify that we selected only non-ambiguous orientations.


We use the subset $\{ {R_i} \in \{R\}, ambiguity(R_i, A) \leq r_0 \}$ to train the classifier on non-ambiguous data.

We first estimate the possible object hypothesis (i.e. $A$ or $B$) and their initial orientation (i.e. $R^{A}_{i_0}$ or $R^{B}_{j_0}$), then we  compute the related ambiguity. If the resulting ambiguity is above the threshold $a$, we move the robot to a more discriminative view.  

We train an autoencoder-based model for each group of ambiguous objects  (e.g., pair of the two flavors boxes in Fig.~\ref{IN.fig.babyfood.active}). The trained network is used both for estimating the rotation of the object, as in \cite{Sundermeyer_2018_ECCV}, and for evaluating the similarity metric $sim$ required to implement Definitions 1 and 2. We search in the space of the other objects' orientations using a similarity metric to compare the rendered orientations. 

\par The robot can now perform the active classification of the object (see. Fig \ref{FIG.pipeline}, inference part). Our input is an RGB image from the robot camera and current robot pose. We first employ the Faster R-CNN~\cite{ren2016faster} object detector to obtain a crop of the image containing the object of interest that we will input to the autoencoder.  In case we have more than one pair of ambiguous objects (e.g., bottles and boxes from Fig. \ref{FIG.objects}), we first have to classify between categories formed by ambiguous pairs (e.g., the first category has both bottles, the second category has both boxes). This task can be accomplished by using a standard classifier. Once the category of interest is identified, the ambiguity of objects within the category depends on the orientation w.r.t. the camera. Hence, if the current view is ambiguous,  we move the robot camera to a more discriminative view. 

As mentioned above, we estimate the next best view based on the current robot pose, object pose, and the set of orientations with calculated ambiguities (see Eq.\eqref{EQ.rank.two}). If the ambiguity of current object orientation is below a given threshold, we can directly apply a classifier trained for this particular category of ambiguous objects.


\section{Ambiguity ranking}
\label{SEC.rank}

This section describes the ambiguity ranking procedure (see Fig.~\ref{FIG.pipeline}). We first train the autoencoder-based similarity metric and then estimate the ambiguity rank using this metric.

\subsection{Autoencoder based view similarity}

\label{SEC.Pipeline.aaa-view-similarity}

In this work we use the AAE for object rotation estimation and, in addition, we  use it to implement the similarity metric $sim$ as per Definition 1:
\begin{equation}
sim(view^A_{R_i}, view^B_{R_j}) := cossim(enc(view^A_{R_i}), enc(view^B_{R_j}))
\label{EQ:aae_sim}
\end{equation}
\par The AAE aims at reconstructing a clean view of the object on white background given a crop of the input image containing the object with background, lighting and occlusion augmentations. The reconstruction occurs by encoding the augmented object view to the latent representation (called embedding) and then decoding the latent representation to the non-augmented object view. To correctly apply the image similarity metric on an object's views, the AAE's latent space must be shared among all similar objects. For this purpose, we trained a single autoencoder to reconstruct all the ambiguous objects, \ref{IN.fig.babyfood.active}). We later refer to this autoencoder as \emph{joint AAE}, as views from all ambiguous objects of the same group are jointly used to train the AAE.


We sample object views on a quasi-uniform Fibonacci grid on the sphere around the object and uniform steps by in-plane rotation.%

Having trained the joint AAE, we  can  generate a codebook for every ambiguous object of interest (see Sec. \ref{SEC.Background.Autoencoder}).
\subsection{Similarity based view ranking}
\label{SEC.rank}

\begin{figure*}[ht!]
    \centering
    \begin{subfigure}[t]{0.64\columnwidth}
        \centering
\includegraphics[width=0.99\columnwidth]{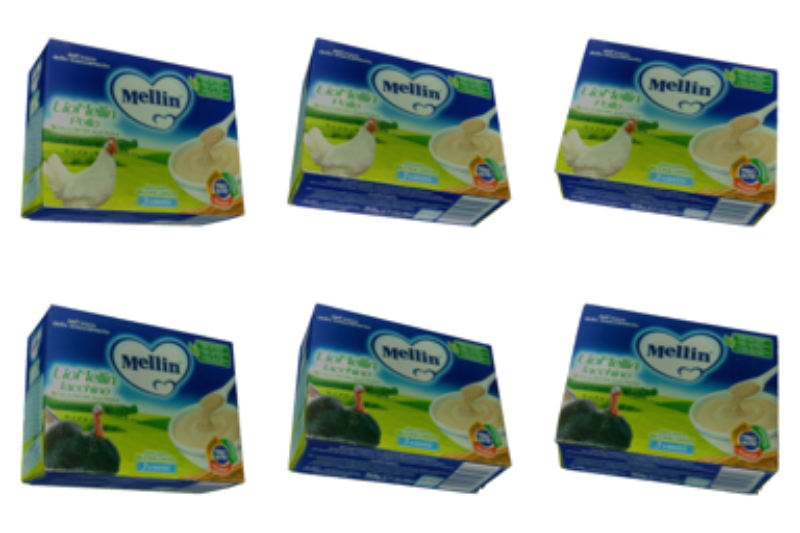}
        \caption{Three most discriminating view pairs of two objects. The main difference between images, a picture of a bird, is seen clearly and occupies a non-negligible part of the image.}
         \label{IN.fig.tops.top}
    \end{subfigure}%
    ~
    \begin{subfigure}[t]{0.64\columnwidth}
\includegraphics[width=0.99\columnwidth]{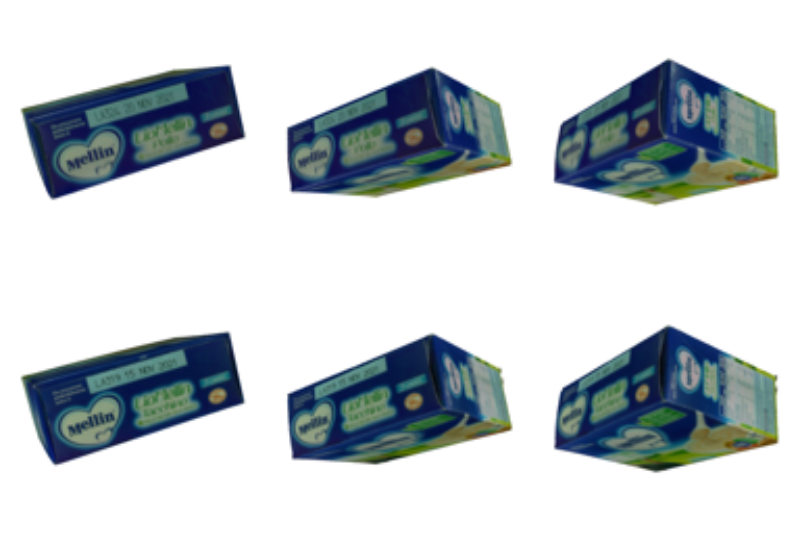}
   \caption{Views from the collection sorted by similarity. Here we see views that lack distinguishable features and views that contain unique features, although in a small part of the view (the picture of a chicken or turkey).}
            \label{IN.fig.tops.mix}
    \end{subfigure}
    ~
    \begin{subfigure}[t]{0.64\columnwidth}
        \centering
\includegraphics[width=0.99\columnwidth]{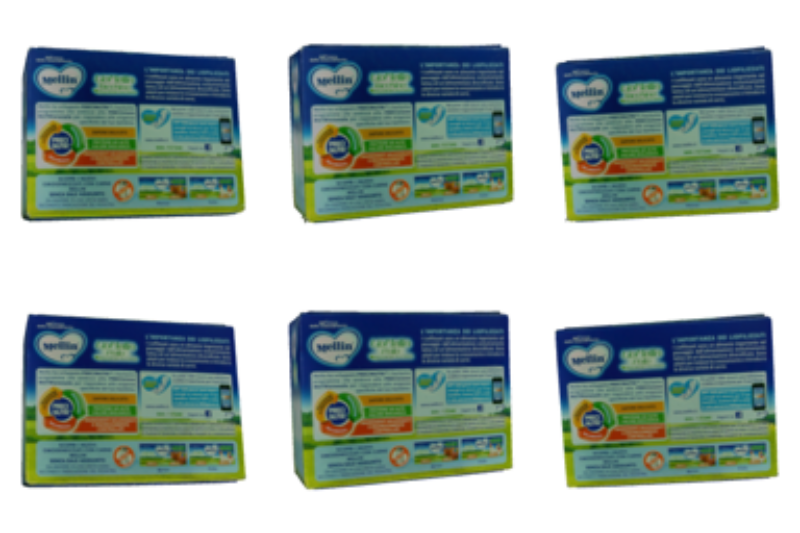}
        \caption{The most similar views as ranked by the trained AAE similarity metric. }
         \label{IN.fig.tops.bottom}
    \end{subfigure}%
    \caption{Visualization of the ranking procedure, sorted views from one object (top row) along with the most similar view from the other object (bottom row). It helps to check if we  performed the view matching and ranking correctly.}
     \label{IN.fig.tops}
\end{figure*}

\label{SEC.Pipeline.view-ranking}
In general, most similar view pairs might have independent orientations. Therefore, to estimate the initial ambiguity of the orientiation of the object $A$ (i.e. $R^A_0$), we find the most similar view $view_{R_x}^B$ from the other object $B$ by maximizing the term 
\begin{equation} 
max_{R} sim(view^A_{R_0}, view^B_{R}) 
\label{EQ:closest_view_min}
\end{equation} 
using the trained similarity metric $sim$ as per Eq. \eqref{EQ:aae_sim}.

This procedure might result computationally expensive. Thus, we sample fewer views over the object compared to those considered while generating the codebook. Note also that here we do not need to sample orientations that differ only in rotations round the camera's optical axis as the resulting images are identical up to a rotation.

\begin{figure}[h]
\centering 
\begin{subfigure}[t]{0.6\columnwidth}
        \centering
\resizebox{0.99\textwidth}{!}{
\begingroup%
  \makeatletter%
  \providecommand\color[2][]{%
    \errmessage{(Inkscape) Color is used for the text in Inkscape, but the package 'color.sty' is not loaded}%
    \renewcommand\color[2][]{}%
  }%
  \providecommand\transparent[1]{%
    \errmessage{(Inkscape) Transparency is used (non-zero) for the text in Inkscape, but the package 'transparent.sty' is not loaded}%
    \renewcommand\transparent[1]{}%
  }%
  \providecommand\rotatebox[2]{#2}%
  \newcommand*\fsize{\dimexpr\f@size pt\relax}%
  \newcommand*\lineheight[1]{\fontsize{\fsize}{#1\fsize}\selectfont}%
  \ifx\svgwidth\undefined%
    \setlength{\unitlength}{221.02910571bp}%
    \ifx\svgscale\undefined%
      \relax%
    \else%
      \setlength{\unitlength}{\unitlength * \real{\svgscale}}%
    \fi%
  \else%
    \setlength{\unitlength}{\svgwidth}%
  \fi%
  \global\let\svgwidth\undefined%
  \global\let\svgscale\undefined%
  \makeatother%
  \begin{picture}(1,0.77350695)%
    \lineheight{1}%
    \setlength\tabcolsep{0pt}%
    \put(0,0){\includegraphics[width=\unitlength,page=1]{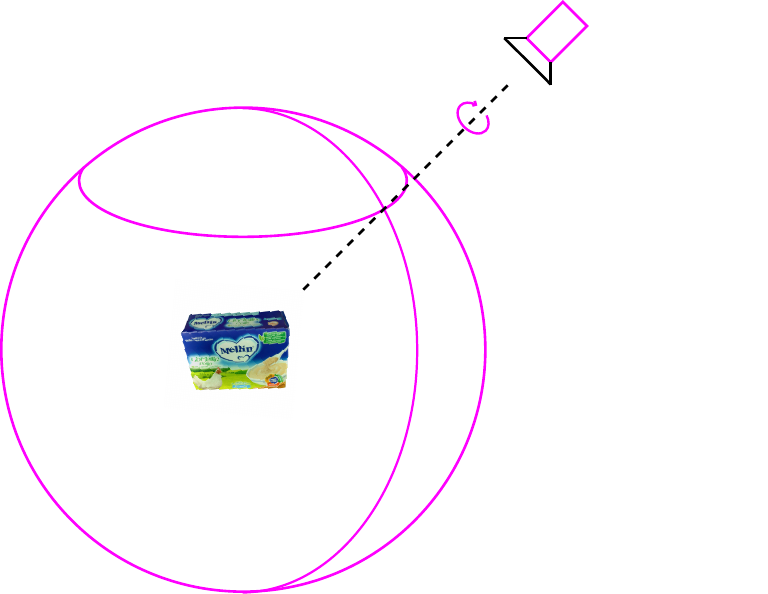}}%
    \put(0.6176444,0.56887488){\color[rgb]{0,0,0}\makebox(0,0)[lt]{\lineheight{1.25}\smash{\begin{tabular}[t]{l}in-plane rotation axis\end{tabular}}}}%
    \put(0.2965006,0.12896841){\color[rgb]{0,0,0}\makebox(0,0)[lt]{\lineheight{1.25}\smash{\begin{tabular}[t]{l}$\phi = const$\end{tabular}}}}%
    \put(0.22101912,0.47890661){\color[rgb]{0,0,0}\makebox(0,0)[lt]{\lineheight{1.25}\smash{\begin{tabular}[t]{l}$\theta = const$\end{tabular}}}}%
  \end{picture}%
\endgroup%

}
\caption{$\theta$, $\phi$, and in-plane axis w.r.t to object.}
\end{subfigure}~
\begin{subfigure}[t]{0.39\columnwidth}
    \centering
	\includegraphics[width=0.9\columnwidth]{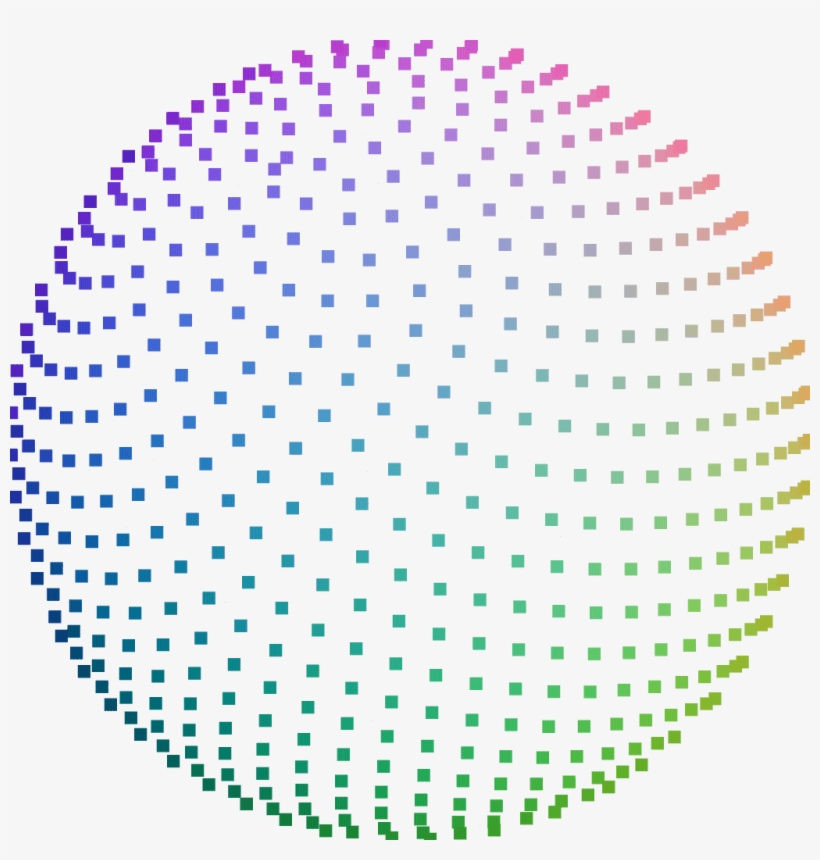}
	\caption{Fibonnaci pseudo-uniform sphere grid example.}
    \label{FIG.sphere}
\end{subfigure}
\caption{Fibonacci pseudo-uniform sphere grid.}
\vspace{-1em}
\end{figure}
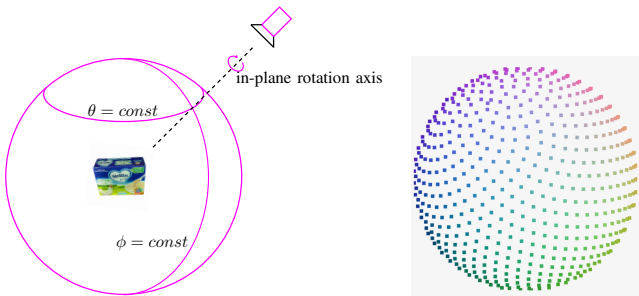

 We replace the cropped image (Fig. \ref{FIG.AAE.training}) with the $view^A_{R}$ and take the codebook from the object $B$. In order to find the most similar view $view_{R_x}^B$, we perform a descent search in the space of the rotations, parametrized as Euler angles, in order to maximize the term in Eq. \eqref{EQ:closest_view_min}. 

We remark that the AAE similarity metric adopted in Eq. \eqref{EQ:closest_view_min} was originally designed to estimate the 3D orientation of an object of interest, hence the devised procedure produces correctly aligned view pairs, as can be seen in Fig.~\ref{IN.fig.tops}.

The output of the matching procedure is a list of tuples $(sim(R^A_i, R^B_{j}), R^A_i, R^B_{j})$, sorted by similarity metric. A useful feature of this approach is that we can render view pairs $(view^A_{R_i}, view^B_{R_{j}})$ and \emph{visually check the correctness} of matching and sorting by ambiguity (see Fig. \ref{IN.fig.tops}).
\section{Training classifiers}
In this section, we address Question 1. We fist show how to split the dataset, then we show how to classify ambiguous objects. To avoid overfitting we validate the classifiers and select the threshold based on a dataset of real images.
\subsection{Splitting the dataset}
\label{SEC.Pipeline.split}
Once we sorted object orientations $\{R\}_{train}$ by their ambiguity, we have to select an ambiguity threshold $a$ such that $ambiguity(view_{R_i}) < a, R_i \in {R}_{dataset}$. This way, the classifier can be trained only on non-ambiguous views. However, we found challenging to select the ambiguity thresholds given only synthetic images rendered from the 3D model. By plotting the ambiguity for the set of sorted rotations in $\{R\}_{train}$, we observed there is no sharp change of ambiguity that could indicate a possible ambiguity threshold (e.g., see AAE baseline on Fig. \ref{COMP.fig}). 
Moreover, we found that the training procedure can be easily compromised by the imperfections in the 3D models, causing overfitting. Therefore, we compute the accuracy on a small dataset of real objects, which then helps us in defining the ambiguity threshold.

%

\subsection{Classification of ambiguous objects and between the categories of ambiguous objects}
In this work, we mainly focus on a classification \emph{inside} a pair or a group of ambiguous objects. However, we might face a situation when we have more than one category of ambiguous objects and some other non-ambiguous objects. 
\label{SEC.Pipeline.in-group-classification}

\section{Inference}
In this section, we address Question 2. First, we need to determine if an acquired view is ambiguous. In that case, we have to find the camera pose where we expect to observe the least ambiguous view taking into account robot motion constraints. However, after this additional movement, the new view may still be more ambiguous than expected, for example due to errors in the estimation of the object orientation. In general, we cannot expect to move the robot to the best configuration in one step, and several movements could be required to converge to a configuration that is suitable for the classification of the object identity. The overall process consists of several stages that we report hereafter:
\paragraph{Object detection}
We first crop the image as both the AAE and our classifier 
 require a square crop with the object in the center.
\paragraph{Classification between categories of ambiguous objects}
In the case of several groups of ambiguous objects, we must first identify in which group a given object falls. As we trained different autoencoder weights for each group, this step of classification provides us with the correct choice of weights for both the joint autoencoder and the in-group classifiers. 
\paragraph{Object orientation and view ambiguity estimation}
To estimate the view ambiguity, we need to first estimate the object orientation. Hence, we need to pass the image through the encoder of the AAE to get the associated embedding and maximize the similarity between this embedding and the object codebook that has been evaluated offline (see Sec. \ref{SEC.Background.Autoencoder}). As we have at least two in-group classes, each with an associated codebook to be used to estimate the orientation of the object (see Sec. \ref{SEC.Background.Autoencoder}), we may get different ambiguity ranks given that they depend on the orientation and on the in-group object class. If the mean of these ambiguity ranks is below the ambiguity threshold (see Sec. \ref{SEC.Pipeline.aaa-view-similarity}), we apply an in-group classifier and return the result. Otherwise, we have to move the robot camera to an appropriate pose.
\paragraph{Next best view for object classification} 
The next view of the object must be such that the classification ambibuity is at least reduced. However, the robot must deal with  reachability constraints and avoid collisions with the surrounding objects. Therefore, the possible view orientations are limited to the set $\{R\}_{reachable}$ satisfying these constraints. Hence, we compute the next best camera-to-object orientation $R^{best}_{next}$ with the following minimization problem:

\begin{equation}
\begin{split}
&R^{best}_{next} = arg\min_{R_{next}} {\sum_{i,j} rank(R_{next}^T R^{A_j}_{obj_i})} \\
&subject~to: R_{next} \in \{R\}_{reachable}
\end{split}
\label{EQ.next-best-view}
\end{equation}

Here we used the mean to calculate the combined ambiguity rank of expected views assuming different object classes $A_j$ and orientations $R^{A_j}_{obj_{i}}$. Having found the next best robot pose, we move the robot, get the new camera input, and repeat the inference steps. We terminate the robot movements when we achieved a view with an ambiguity lower than the desired threshold or when the current view is least ambiguous among the possible robot poses ($R^{best}_{next}$ equals to current robot orientation). 

\section{Comparison against other feature extraction and matching methods}


In this section, we compare our autoencoder-based similarity with other metrics. We evaluated similarity metrics on pairs $(view^A_{R_i}, view^B_{R_{j}})$, where the second view in each pair is the most similar to the first one according to the AAE-based image similarity. 

\begin{figure}[h]
    \centering
    \begin{subfigure}[t]{0.49\columnwidth}
        \centering
\includegraphics[width=0.99\columnwidth]{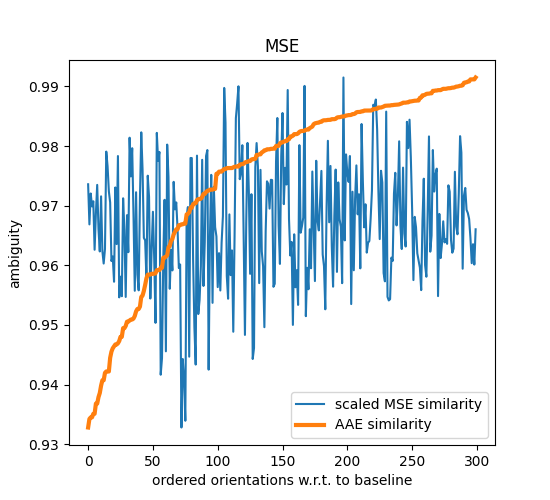}

         \caption{Mean squared error}
         \label{COMP.fig.p2p}
    \end{subfigure}%
	~    
    \begin{subfigure}[t]{0.49\columnwidth}
        \centering
\includegraphics[width=0.99\columnwidth]{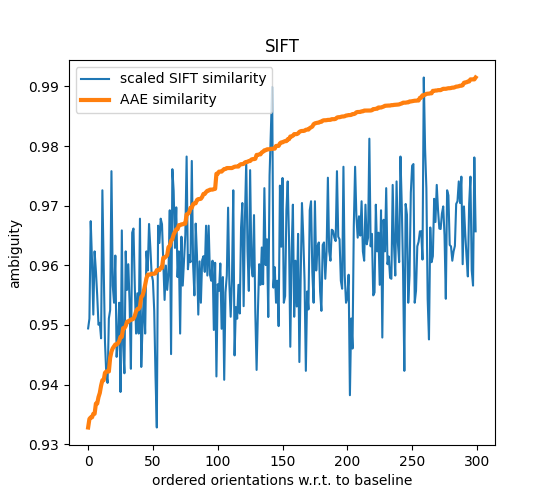}

    \caption{ SIFT descriptors} 
    \label{COMP.fig.sift}
    \end{subfigure}%
    \\
    \begin{subfigure}[t]{0.49\columnwidth}
\includegraphics[width=0.99\columnwidth]{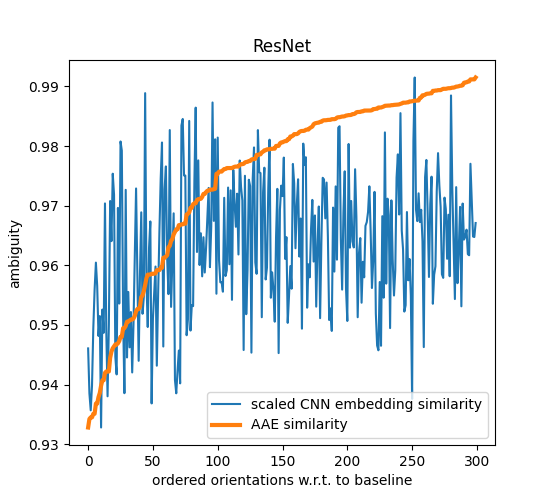}

		\caption{ResNet embedding}
		\label{COMP.fig.resnet}
	\end{subfigure}%
	~
    \begin{subfigure}[t]{0.49\columnwidth}
\includegraphics[width=0.99\columnwidth]{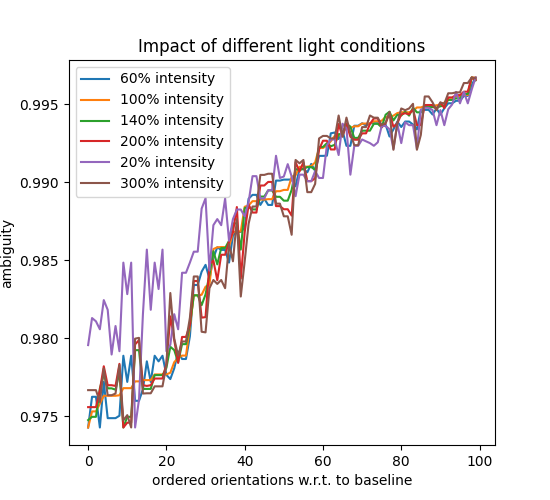}

		\caption{Different light conditions}
		\label{COMP.fig.light}
	\end{subfigure}%
	\caption{Different similarity metrics w.r.t. to AAE baseline. X axis shows indexes of view pairs sorted by AAE ambiguity rank. Y axis shows scaled similarity metric for the pair indexed by X.}
    \label{COMP.fig}
\end{figure}

%

\begin{figure*}[h]
    \centering
    \begin{subfigure}[t]{0.7\columnwidth}
        \centering
\includegraphics[width=0.99\columnwidth]{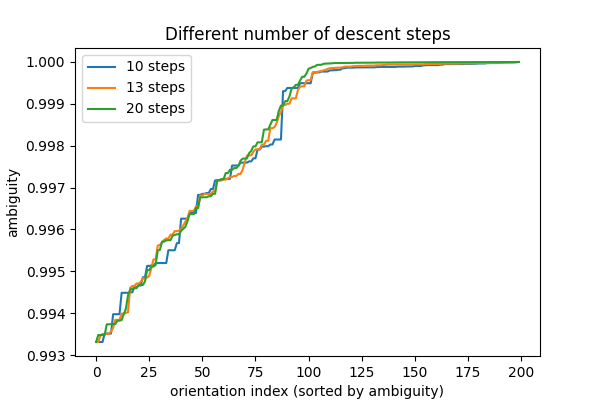}
         
         \caption{A ambiguity plot for emulated mustard bottles pair. Three plots were acquired with the different number of steps in the descent search part of ambiguity ranking. A higher number of steps makes the ambiguity rank approach one and results in a smoother plot.  
         }
         \label{EXP.fig.ideal.plot}
    \end{subfigure}%
    ~
    \begin{subfigure}[t]{0.41\columnwidth}
\includegraphics[width=0.99\columnwidth]{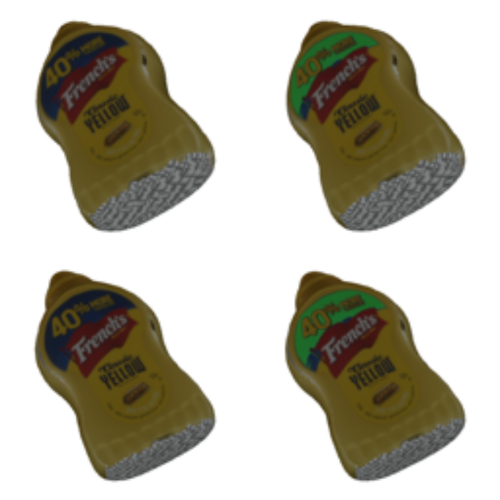}
            
            \caption{Example of views that have low ambiguity. The modified part of the texture is visible and covers a significant part of the object view.}
            \label{EXP.fig.ideal.top}
    \end{subfigure} 
    ~
    \centering
    \begin{subfigure}[t]{0.41\columnwidth}
        \centering
\includegraphics[width=0.99\columnwidth]{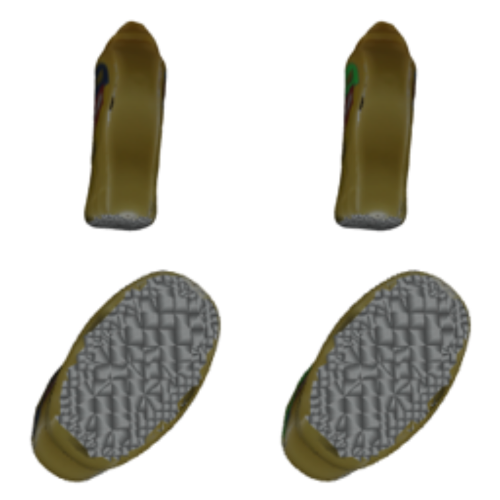}
         
            \caption{Example of views that have ambiguity close to the threshold. The views are ambiguous, as they only differ in a small part of the view. }
            \label{EXP.fig.ideal.mid}
    \end{subfigure}%
    ~
    \begin{subfigure}[t]{0.41\columnwidth}
\includegraphics[width=0.99\columnwidth]{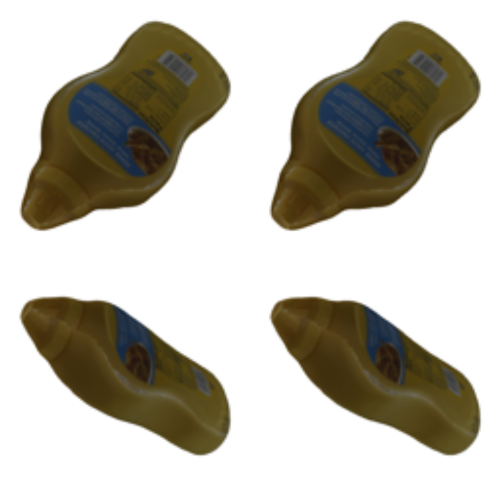}
            
            \caption{Example of ambiguous views. We see unmodified part of the object therefore similarity approaches one. }
            \label{EXP.fig.ideal.bot}
    \end{subfigure}\\
\caption{Experiments with YCB mustard bottle and the same bottle with modified texture}
\label{EXP.fig.ideal}
\vspace{-1em}
\end{figure*}

The first metric we compare to is the pixelwise mean squared error (MSE) between the images of the two  views (Fig. \ref{COMP.fig.p2p}). The second metric is based on a comparison between the SIFT descriptors \cite{lowe2004distinctive} evaluated on the two images (Fig. \ref{COMP.fig.sift}). The third metric is the cosine similarity between the embeddings extracted from the two images using a feature extractor based on a CNN. Specifically, we adopted the ResNet-50 architecture from which the last fully connected layer has been removed (see Fig. \ref{COMP.fig.resnet}).
We observed that the plots corresponding to the three metrics barely correlate to the AAE baseline (see Fig. \ref{COMP.fig}), this means that these metrics are unable to discriminate ambiguous and non-ambiguous views.
\par In addition, we checked how different light conditions affect ambiguity discrimination. To this end, we rendered the views while varying the light intensity. For a broad range of configurations of the light intensity, the ability to discriminate between the two objects seem mostly unaffected (see Fig. \ref{COMP.fig.light}).

\label{SEC.comparison}
\section{Experiments}
\label{SEC.Experiments}
In this section, we present the validation of our approach with a set of experiments, both in simulation and on a real robot. 
We implemented the whole pipeline for training and inference  as a Python package, available online\footnote{https://github.com/safoex/aoc}. The package includes an implementation of autoencoder-based similarity metric using the PyTorch Lightning~\cite{falcon2019pytorch} framework while following the original work on AAE including augmentations \cite{Sundermeyer_2018_ECCV}. We fine-tune the ResNet-18 CNN classifier pre-trained on the ImageNet dataset.
For object detection, we use a pre-trained Faster R-CNN network from the Torchvision package \cite{torch}.

\subsection{Simulated experiment}
\label{SEC.Experiments.ideal}

As mentioned earlier in this work, existing datasets do not contain pairs of objects with substantial ambiguity. For this reason, we generated data featuring pairs of ambiguous objects using the publicly available 3D model of the ``mustard bottle'' object from the YCB model set~\cite{ycb}. We used both the original 3D mesh of the object and a modified version of it where we introduced a variant in the texture of the front side. The resulting pair of objects has some distinguishing features on the front side and several completely identical views (see Fig.~\ref{EXP.fig.ideal.top}). The ambiguity rank of such identical views must be exactly $1$. Fig. \ref{EXP.fig.ideal.plot} shows the obtained ambiguity plot. The ambiguity approaches one for half of the object orientations (half of the sorted views on the Fig \ref{EXP.fig.ideal.plot}), therefore, a value close to one should be used as an ambiguity threshold for this pair of objects. Fig. \ref{EXP.fig.ideal} shows examples from the sorted view pairs. The lowest-ranked views look visually the most distinguishable as the view shows the modified texture (Fig. \ref{EXP.fig.ideal.top}) while the views with higher ambiguity do not feature the modified texture (Fig. \ref{EXP.fig.ideal.bot}) (this is also corroborated by the shape of the plot in Fig. \ref{EXP.fig.ideal.plot}). Near the middle of the sorted views, where ambiguity approaches one, we see that differences between two views are very small (Fig. \ref{EXP.fig.ideal.mid}). Hence, in this case, we can choose a similarity threshold around $a \simeq 1-5e^{-4}$. Only on view with the ambiguity less than $a$ would we see the non-ambiguous feature. The proposed tests in simulation show that using the autoencoder to rank the ambiguity of the views is sound, as formulated in Section~\ref{SEC.Overview}.
\subsection{Experimental setup}
\begin{figure}[b]
\includegraphics[width=0.96\columnwidth]{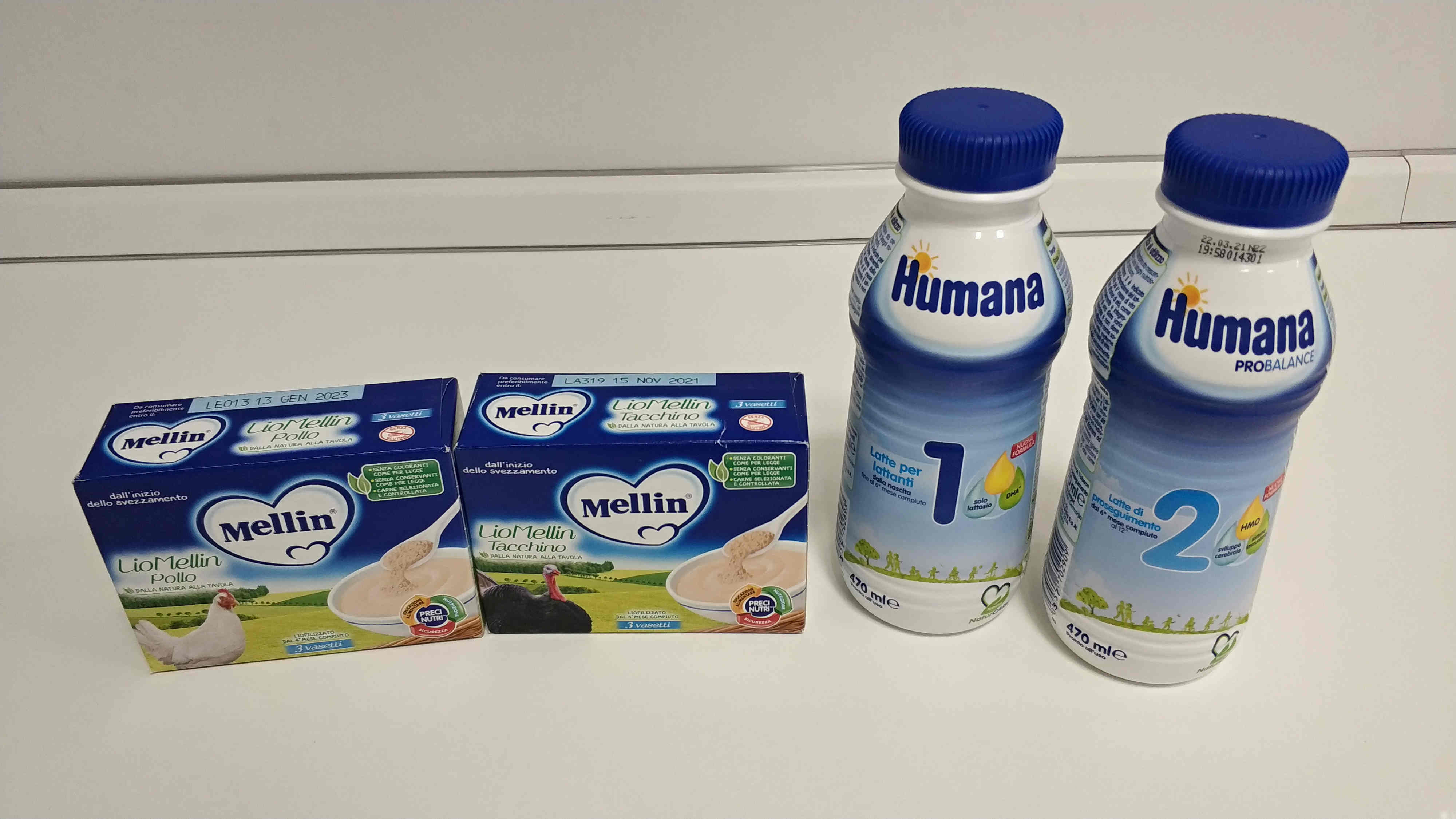}
        \caption{Two groups of ambiguous objects that were used in the experiments.
        \label{FIG.objects}
        }
\end{figure}

Despite the lack of ambiguous objects in the available datasets, pairs of ambiguous objects could be easily found in the real world, e.g., among packaged food available in supermarkets. For our experiments, we chose two groups of ambiguous objects. One of them is a pair of food boxes, the other is a pair of bottles (see Fig. \ref{FIG.objects}). We retrieved the 3D models using a commercially available 3D scanner, namely the \emph{Shining 3D} scanner. Although the reconstructed models had small artifacts, the quality of the models resulted good enough for our purposes. We used a Franka Emika Panda robot with an Intel RealSense D415 camera, mounted on the end-effector, to acquire image views. 

\subsection{Classification test on recorded data}
\label{SEC.Experiments.recorded}

In a first set of experiments, we acquired a set images using the camera mounted on the robot in several fixed poses. The poses were chosen such that the object appeared in the center of the image plane. Furthermore, we assumed to know the Cartesian position of the object with respect to the robot reference frame. This constraint does not affect the verification of the pipeline but simplifies the setup and the analysis. 

We recorded data by following a fixed trajectory along several parallel circles on a sphere centered on the Cartesian position of the object. Each trajectory was parametrized in terms of Azimuth $\phi$ and elevation angles $\theta$. For each value of the elevation, we sampled uniformly along the space of Azimuth angles (see Fig.\ref{FIG.sphere}). Poses of the end-effector that were not reachable, due to constraints in the joints configuration, were not considered while acquiring the data.
The collected data consisted of the images captured from the robot camera along with the relative pose between the camera and the object. We collected a small dataset of object views from different sides, including ambiguous and non-ambiguous views. The trajectory consisted of three parallel circles (i.e., with a constant elevation $\theta$) on the sphere (see Fig.\ref{FIG.sphere}). We sampled $128$ points each parallel circle. We repeated the execution four times for each object, each time considering a different orientation of the object with respect to the robot root frame.

%
\begin{figure}[b]
\includegraphics[width=0.96\columnwidth]{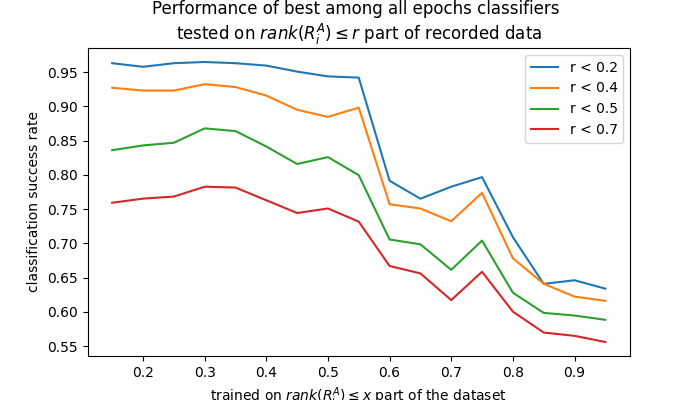}
        \caption{The performance of best classifier for each training ambiguity threshold used (X axis) and for different part of training data (less than certain ambiguity rank, see legend). 
        \label{FIG.Experiments.recorded.sumup}
        }

\end{figure}

We then trained the object classifiers on the collected data while varying the threshold that separates ambiguous views from non-ambiguous views.  Fig. \ref{FIG.Experiments.recorded.sumup} shows the results in terms of classification accuracy. The classification performance drops significantly when the training data include object orientations with ambiguity higher than $0.5$. The objects used in the real experiment have visually distinguishable features from one side only. We found better classification performance for the non-ambiguous part of the object's views than the whole object. This is clear also from Fig. \ref{FIG.Experiments.recorded.sumup}, as plots for evaluation on more ambiguous parts of recorded data are under the plot for lower ambiguity of test data. Moreover, classifiers trained on both ambiguous and non-ambiguous orientations tend to predict one class with high certainty irrespective of the actual identify within the group of ambiguous objects.
\subsection{Active classification results}
In a second set of experiments, we tested the active classification capabilities of the proposed pipeline. We made two types of experiments. First, as we had recorded data for many camera-to-object orientations, we performed an \emph{offline} active vision test constraining robot movements $\{R\}_{reachable}$ by recorded positions. We compared this offline experiment against the random baseline when the robot picked the next view randomly from $\{R\}_{reachable}$.  Next, we performed an \emph{online} active vision test, where the robot starts from a random pose, hence a random orientation of the camera, and then tries to reach the pose where the ambiguity rank of the current object view is less than $0.4$ and classify the object.
\par To perform the offline active vision experiments, we recorded images using a similar approach as above but with a radius of $0.3m$, $32$ steps on $5$ parallel circles on the sphere around the object. We compared the next best view selection against the random next view. In all cases, we outperform random next view selection (see Fig. \ref{EXP.fig.av}). Note that as bottles have a symmetrical shape, they are more challenging for orientation estimation and require additional views.

\begin{figure}[t]
    \centering
    \begin{subfigure}[t]{0.49\columnwidth}
        \centering
\includegraphics[width=0.99\columnwidth]{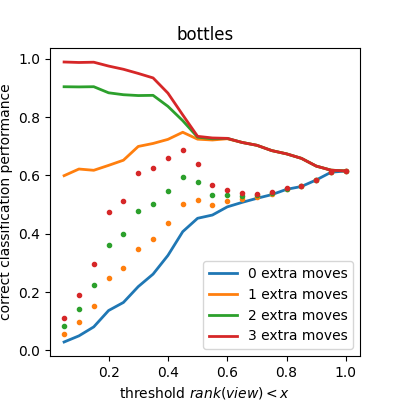}
         \caption{Plots for the bottles (see Fig. \ref{FIG.objects}). }
         \label{EXP.fig.off_av_bottles}
    \end{subfigure}%
    ~
    \begin{subfigure}[t]{0.49\columnwidth}
\includegraphics[width=0.99\columnwidth]{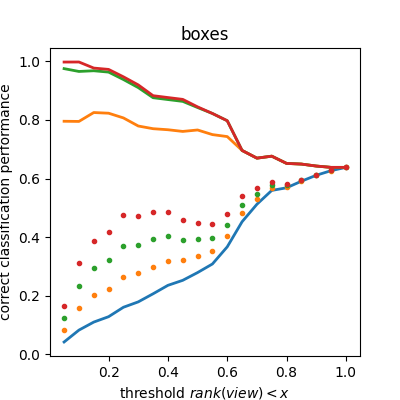}
            \caption{Plots for the boxes (see Fig. \ref{FIG.objects}). }
            \label{EXP.fig.off_av_boxes}
    \end{subfigure}
    \caption{Classification success probability in offline active vision experiment. X axis represents the ambiguity rank threshold chosen to terminate the active perception. Different plots on the same figure represent the classification performance for different numbers of extra robot movements allowed. Dashed represents the performance of the random baseline. For example, red (highest) plot on the Fig. \ref{EXP.fig.off_av_bottles} shows the fraction of correct active classifications after no more than three extra robot movements.}
    \label{EXP.fig.av}
\end{figure}
\begin{figure}[t]
    \centering
    \begin{subfigure}[t]{0.49\columnwidth}
        \centering
\includegraphics[width=0.99\columnwidth]{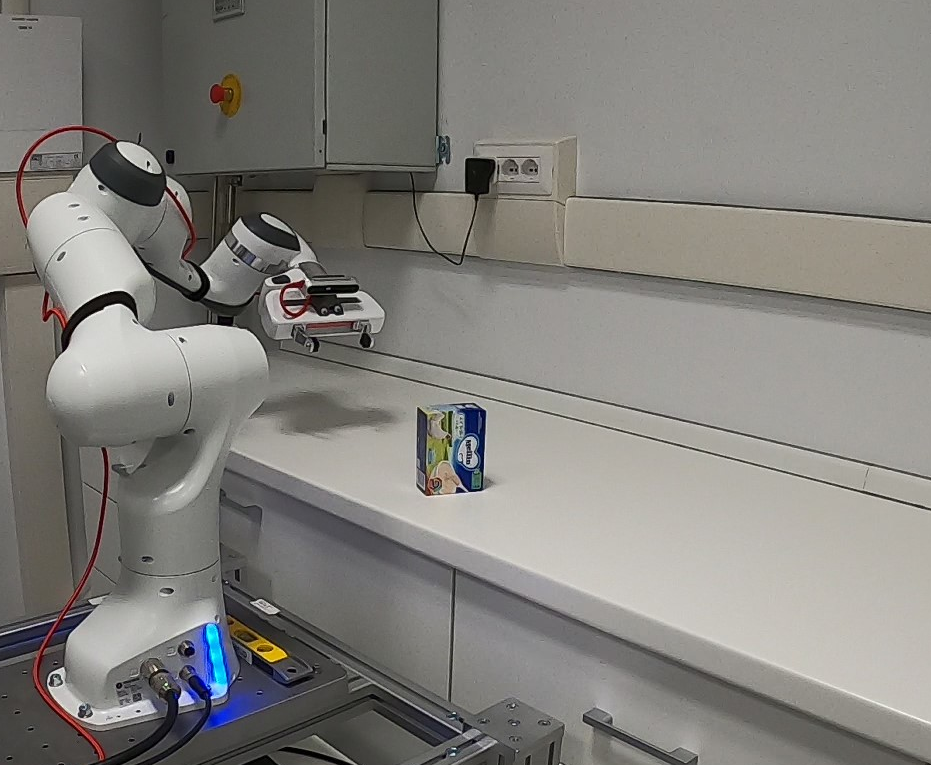}
         \caption{Initial pose of the robot.}
         \label{EXP.fig.av1}
    \end{subfigure}%
    ~
    \begin{subfigure}[t]{0.49\columnwidth}
\includegraphics[width=0.99\columnwidth]{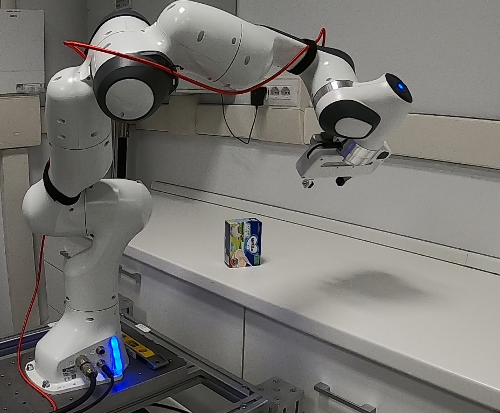}
            \caption{Robot pose after the next best view selection. }
            \label{EXP.fig.av2}
    \end{subfigure}
    \caption{An example of robot movement to improve classification. The robot moves its viewpoint from the back, ambiguous,  view of the ``chicken box'' (from which classification is impossible), to the front, where the box has discriminative features.}    
    \label{EXP.fig.av}
\end{figure}

Regarding active vision experiments, each experiment started from a random camera-to-object pose and terminated by reaching the ambiguity threshold or when the current robot pose was optimal, according to Eq. \ref{EQ.next-best-view}. Overall, $230$ experiments were performed, of which only 20 finished with incorrect classification results, corresponding to a classification accuracy of about $91\%$. In these experiments, we set the ambiguity threshold to $0.4$ i.e.,  if the robot acquired an image that was evaluated with a rank less than $0.4$, we stopped the classification  procedure. The obtained results correlate with those obtained on prerecorded data, and described in Sec. \ref{SEC.Experiments.recorded}, when considering the same ambiguity rank of  $0.4$ (see Fig. \ref{FIG.Experiments.recorded.sumup}). An example of the action recognition task is shown in Fig. \ref{EXP.fig.av}. 

\label{SEC.Experiments.active}
\section{Conclusions}
This paper argued that everyday objects have ambiguous views that make standard classification approaches challenging to apply. We propose a novel active perception strategy based on view ambiguity estimation employing an autoencoder embedding. We validated our approach on a real robot using household objects, demonstrating its feasibility and performance. 

In future work we plan to investigate the use of the autoencoder similarity metric to cluster groups of ambiguous objects within a given dataset, and extend this work for selecting best views to perform object discrimination in presence of occlusions. Another direction of research is  to employ the same similarity metric for handling symmetries in pose estimation.

\section{Acknowledgements}
We thank Fabrizio Bottarel for installing the Franka Emika Panda robot and the associated  software ecosystem. This work was supported by the European H2020 project No. 730994 (TERRINet) and ERA-NET CHIST-ERA call 2017 project HEAP.
    \bibliography{root}

    \bibliographystyle{ieeetr}
\end{document}